\documentclass[lettersize, journal]{IEEEtran}
\usepackage[T1]{fontenc}
\usepackage{amsmath,amsfonts}
\usepackage{algorithmic}
\usepackage{algorithm}
\usepackage{array}
\usepackage{textcomp}
\usepackage{url}
\usepackage{balance}
\usepackage{verbatim}
\usepackage{graphicx}
\usepackage{cite}
\usepackage{stfloats}
\usepackage{booktabs}
\usepackage{color}
\usepackage[dvipsnames]{xcolor}
\usepackage{dsfont}
\usepackage{colortbl}
\usepackage{bm}
\definecolor{DodgerBlue}{HTML}{1E90FF}
\usepackage[urlcolor=DodgerBlue, 
citecolor=DodgerBlue, 
linkcolor=DodgerBlue, 
colorlinks=true,]{hyperref}
\usepackage{multirow}
\usepackage{soul}
\usepackage{ragged2e}

\usepackage{multicol}
\usepackage{makecell}
\usepackage{booktabs}
\usepackage{amssymb}
\usepackage{bbding}
\usepackage{pifont}
\usepackage{wasysym}
\begin{document}

\title{Duawlfin: A Drone with Unified Actuation for Wheeled Locomotion and Flight Operation}

\author{Jerry Tang, Ruiqi Zhang, Kaan Beyduz, Yiwei Jiang, Cody Wiebe, Haoyu Zhang, \\Osaruese Asoro, and Mark W. Mueller 
\thanks{The authors are with the Department of Mechanical Engineering, University of California Berkeley, Berkeley, CA 94720, USA.}
\thanks{Correspond to: Jerry Tang and Ruiqi Zhang. E-Mail: {\tt{\small jerrytang, richzhang@berkeley.edu}}}
}


\maketitle
\begin{abstract}
This paper presents Duawlfin, a drone with unified actuation for wheeled locomotion and flight operation that achieves efficient, bidirectional ground mobility. Unlike existing hybrid designs, Duawlfin eliminates the need for additional actuators or propeller-driven ground propulsion by leveraging only its standard quadrotor motors and introducing a differential drivetrain with one-way bearings. This innovation simplifies the mechanical system, significantly reduces energy usage, and prevents the disturbance caused by propellers spinning near the ground, such as dust interference with sensors. Besides, the one-way bearings minimize the power transfer from motors to propellers in the ground mode, which enables the vehicle to operate safely near humans. We provide a detailed mechanical design, present control strategies for rapid and smooth mode transitions, and validate the concept through extensive experimental testing. Flight-mode tests confirm stable aerial performance comparable to conventional quadcopters, while ground-mode experiments demonstrate efficient slope climbing (up to 30°) and agile turning maneuvers approaching 1g lateral acceleration. The seamless transitions between aerial and ground modes further underscore the practicality and effectiveness of our approach for applications like urban logistics and indoor navigation. All the materials including 3-D model files, demonstration video and other assets are open-sourced at {\tt\small\url{https://sites.google.com/view/Duawlfin}}.
\end{abstract}


\section{Introduction}
\subsection{Background}

\IEEEPARstart{U}{nmanned} Aerial Vehicles~(UAVs) exhibit notable advantages in terms of mobility and terrain adaptability. Due to their ability to operate in three-dimensional space~\cite{mark2022design, thusoo2021quadrotors}, UAVs possess superior maneuverability can traverse complex or unstructured environments—such as rugged terrain, dense vegetation, or urban obstacles—without the constraints imposed by ground contact. Such mobility makes UAVs especially suitable for tasks that require rapid deployment, flexible routing and access to otherwise inaccessible locations, like delivery, remote sensing, and infrastructure inspection and monitoring~\cite{samouh2020multimodal, raivi2023drone, bridgelall2024remote}.  However, UAVs also suffer from several intrinsic limitations, which primarily stem from the aerodynamic and propulsion requirements associated with sustained flight. For instance, multirotor UAVs must constantly generate lift equal to their weight through high-speed rotation of propellers, resulting in significant energy expenditure even when hovering~\cite{jain2020iros}.
\begin{figure}[h!]
    \centering
    \includegraphics[width=0.85\linewidth]{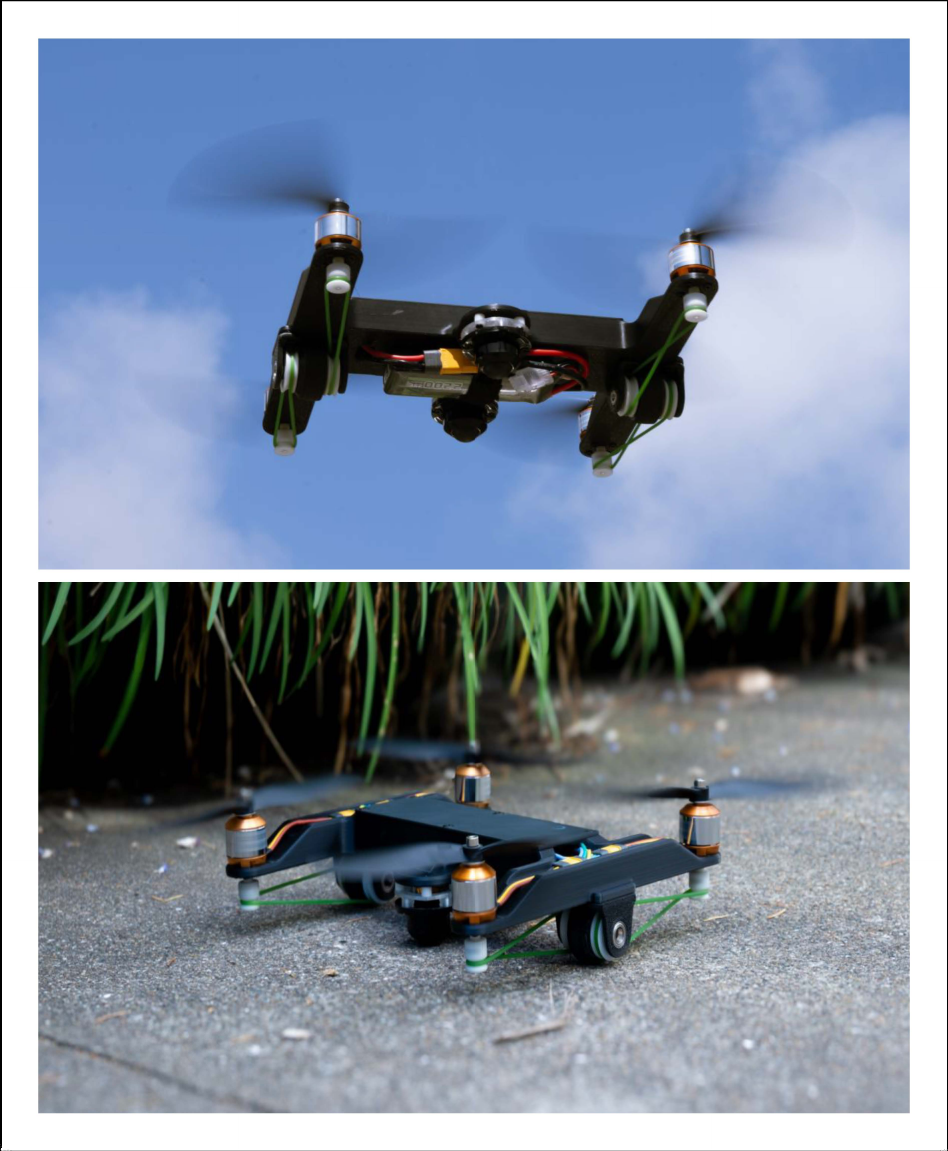}
    \caption{Top: Duawlfin is in aerial mode. Bottom: Duawlfin is runing on the sidewalk in ground mode. The propellers are still turning at low speed since the friction in the one-way bearings' free mode.}
    \label{fig:Duawlfin}
\end{figure}

Ground mobility can alleviate some of these issues. By letting aerial robots roll or drive when airborne travel is not essential, energy consumption can be substantially reduced and safety hazards near the ground can be minimized~\cite{samouh2020multimodal}. However, achieving this aerial–ground versatility is non-trivial. Hybrid robots must incorporate additional hardware or specialized transmissions to enable terrestrial motion, typically adding weight, complexity, and potential failure points~\cite{kalantari2020drivocopter, qin2020hybrid, sugihara2024design}. The goal, then, is to design a hybrid configuration, which can preserve the UAV’s agility and compactness while gaining the energy and functional benefits of ground locomotion.

\subsection{Literature Review}
Numerous hybrid aerial–ground vehicles have been proposed to strike a balance between flight capabilities and terrestrial efficiency. Existing designs can be broadly divided into two primary approaches based on how they achieve ground propulsion.

\subsubsection{Propeller-driven ground propulsion}
These UAV designs employ angled or redirected thrust from their existing propellers to achieve ground locomotion. Page and Pounds (2014) introduced the Quadroller, a quadrotor with low-friction wheels and skateboard steering trucks that rolls farther than it hovers despite minimal mass penalty~\cite{page2014quadroller}. Sabet et al. (2019) designed Rollocopter, a spherical-shell platform that rolls on extreme terrain and flies, balancing power via an energy-aware controller~\cite{sabet2019rollocopter}. Premachandra et al. (2019) retrofitted a quadcopter’s propellers for ground motion and obstacle avoidance without extra actuators~\cite{premachandra2019study}. Sugihara et al. (2024) proposed Delta, a deformable multirotor that rolls with its entire body and seamlessly switches modes~\cite{sugihara2024design}. Pan et al. (2023) demonstrated Skywalker, which uses an omnidirectional wheel and differential-flatness control to achieve up to a speed of 5 m/s and 75.2\% energy savings on the ground~\cite{neng2023skywalker}. Zhang et al. (2023) designed a quadcopter with passive wheel and leveraged nonlinear model predictive control~(MPC) with maximum $3 m/s$ ground velocity~\cite{zhang2023control}. Similarly, Lin et al. (2024) proposed Skater, a bi-copter with passive wheels that leverages vectored thrust and MPC for robust terrain traversal and steering~\cite{lin2024skater}. Qin et al. (2020) integrated a single passive wheel at a UAV’s base, saving 77\% battery during rolling~\cite{qin2020hybrid}.

Such an approach offers simplicity, as no additional actuators or motors are required. However, using propellers for ground propulsion generally requires higher power than dedicated wheel actuation, particularly during sustained or high-force maneuvers. Continuous spinning of propellers near the ground may also stir up dust and debris, potentially interfering with onboard sensors and affecting other system components. Additionally, prolonged ground operation can increase thermal loads on motors, possibly reducing their service life~\cite{saemi2024heat}, and the spinning propellers pose inherent safety concerns.

\subsubsection{Additional actuators for ground motion}
The second category equips the robot with dedicated actuators—usually motors driving wheels or legs. Kalantari et al. (2020) introduced Drivocopter, a UAV with four actuated spherical wheels for long-endurance hybrid mobility and propeller protection~\cite{kalantari2020drivocopter}. Morton and Papanikolopoulos (2017) developed a compact hybrid that transforms between ground and air modes, shielding flight hardware when grounded~\cite{morton2017mobile}. Zhao et al. (2023) introduced SPIDAR, a quadruped with distributed vectorable rotors for both walking and flight~\cite{zhao2023spider}. Sugihara et al. (2023) presented FlyHuman, a humanoid with wheels and a flight unit enabling aerial, legged, and wheeled locomotion under unified control~\cite{sugihara2023flyhuman}. Adarsh and Dharmana (2018) developed MTMUR, an amphibious robot using custom wheels and floats to navigate land, water, and air~\cite{rs2018multi}. Sihite et al. (2023) introduced M4, which repurposes appendages into wheels, thrusters, and legs for eight distinct locomotion modes~\cite{sihite2023m4}.

The advantage of this design lies in the use of two independent sets of actuators, which eliminates concerns about motor overload and allows each actuator to operate efficiently. Moreover, by decoupling the power sources for ground and aerial locomotion, the propellers are not engaged during ground movement, which enhances operational safety. These designs typically demonstrate robust ground maneuverability but at the expense of higher total mass which in turn reduced flight times. The greater complexity also increases cost and maintenance requirements.

\subsection{Contribution}
Despite the diversity of existing hybrid platforms, current designs typically rely either on propeller-driven rolling or on additional terrestrial actuators to enable ground mobility. The field thus remains open to exploring simpler and lighter solutions capable of bidirectional ground travel by leveraging only existing aerial motors, thereby minimizing mechanical complexity and energy use.

This paper introduces Duawlfin (Figure~\ref{fig:Duawlfin}), a drone with unified actuation for wheeled locomotion and flight operation that achieves efficient, bidirectional ground mobility without requiring additional actuators or relying on propeller thrust for terrestrial locomotion. Ground propulsion is achieved using a mechanical drivetrain powered by differential motor speeds, with one-way bearings that decouple propellers during terrestrial operation. This streamlined approach enables stable and energy-efficient terrestrial locomotion, avoids the complexity and mass associated with adding dedicated ground actuators, and preserves the full functionality of conventional quadrotor flight. Specifically, the contributions of this letter are:
\begin{itemize}
    \item The proposed hybrid design achieves bidirectional ground mobility using the drone’s \textit{existing} motors only, requires \textit{no additional actuators} and avoids \textit{propeller-based ground propulsion}.
    
    \item The proposed system integrates a lightweight drivetrain that enables efficient ground driving without energy loss during flight, supports seamless transitions between modes with no reconfiguration, and allows differential motor control for intuitive bidirectional steering on the ground.
    
    \item Comprehensive real-world tests demonstrate stable and agile flight, smooth ground locomotion, and robust aerial-ground transitions.
\end{itemize}

The remainder of this paper is organized as follows: Section~\ref{sec:design} provides a detailed description of the mechanical design of the hybrid robot, highlighting key structural features and the principles enabling dual-mode locomotion. Section~\ref{sec:control} elaborates on the mode-switching control algorithm developed for hybrid mobility. Section~\ref{sec:exp} presents a comprehensive set of real-world experiments, analyses, and evaluations of the system. Finally, Section~\ref{sec:conclusion} concludes the paper and discusses directions for future research.

\section{Design}
\label{sec:design}
\subsection{Design Objectives}
The goal of this work is to develop a hybrid aerial-ground robot capable of both flying and driving using a single unified actuation system. To minimize hardware complexity and weight, the design reuses the drone’s existing motors for both flight and ground propulsion. A key requirement is to ensure that during ground operation, the propellers do not engage or consume energy, thereby avoiding unnecessary aerodynamic drag and power loss. All added components must be as lightweight and as compact as possible.

\subsection{Key Insights}
\begin{figure*}[th]
    \centering
    \includegraphics[width=17cm]{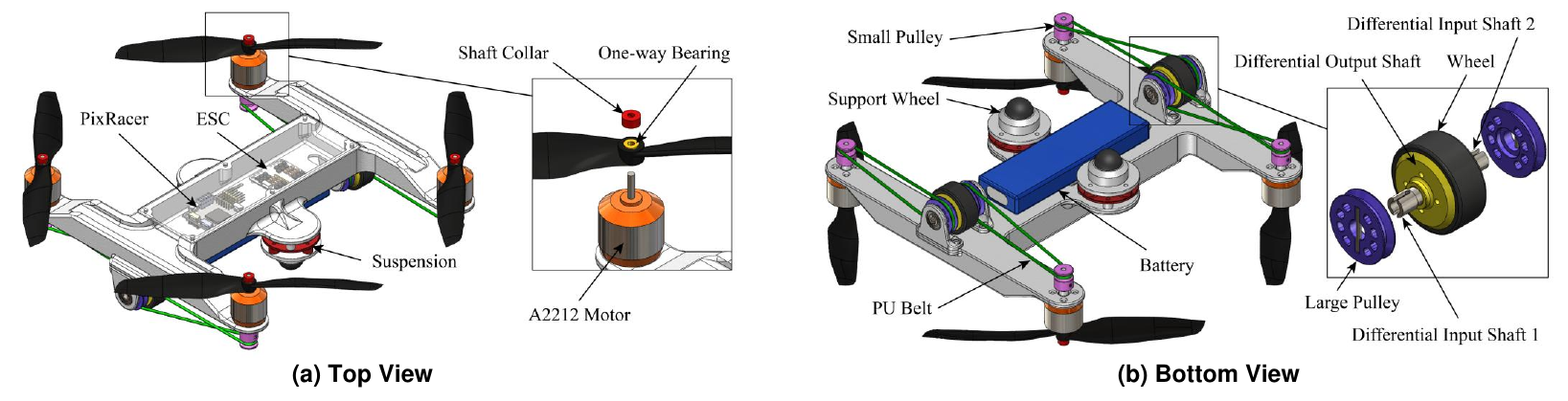}
    \caption{(a): Top-side view of Duawlfin. It shows its aerial propulsion system. The vehicle features four A2212 1400\,kv motors driving 8-inch (8045) propellers. Each propeller is mounted via a one-way bearing and secured by a shaft collar, ensuring that in flight mode the bearings lock to generate thrust, while in ground mode the propellers freewheel. (b): Bottom-side view. It shows the ground drivetrain arrangement. Each motor shaft is equipped with a small pulley that drives a belt connected to a larger pulley on one of the differential’s input shafts. Two opposing motors feed their respective differentials, whose output shafts are directly connected to integrated wheels. Universal ball casters at the front and rear provide stable ground support.}
    \label{fig:cad}
\end{figure*}
Multirotor UAVs typically employ brushless DC motors, which are electrically capable of bidirectional operation but mechanically constrained due to the aerodynamic design of their propellers. Propellers are usually optimized for thrust generation in only one direction, making reverse rotation highly inefficient in terms of thrust-to-power ratio.

This inherent directional asymmetry can be leveraged for passive mechanical decoupling. By mounting each propeller to its shaft through a one-way bearing, it becomes possible to engage the propeller only during forward motor rotation—the optimal direction for aerodynamic thrust—while allowing the propeller to disengage and freely rotate if the motor reverses direction. In such a configuration, reverse motor rotation could potentially be exploited to mechanically harvest torque without unnecessarily spinning the propellers, thus enabling efficient power transfer to a ground drivetrain.

\begin{figure}[h]
  \centering
  \includegraphics[width=\linewidth]{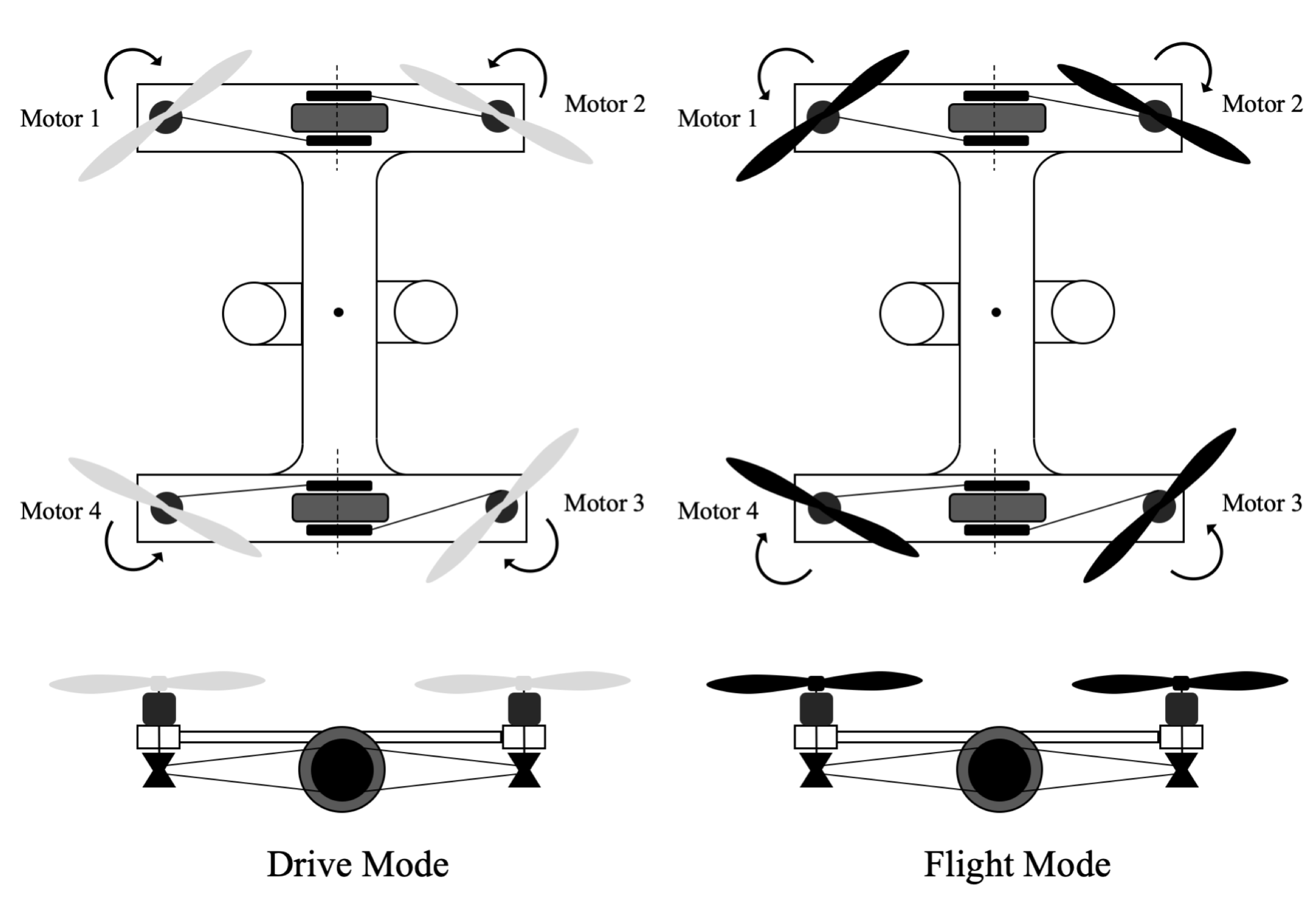}
  \caption{Motor directions for drive and flight modes. In the flight mode, the propellers are engaged with the one-way bearings that allow torques to be transmitted to generate thrust. In the drive mode, the motors reverse and decouple the propellers. The motor speed differences are manipulated to provide full bidirectional ground mobility through the differentials.}
  \label{fig:both_sides}
\end{figure}

Because motors would spin exclusively in reverse during ground mode, directly coupling each motor to a wheel could lead to unidirectional wheel rotation, severely limiting ground mobility. A differential mechanism, commonly used in automotive applications, can resolve this limitation by combining inputs from two opposing motors. By coupling these motors through pulleys to a central differential, wheel rotation becomes dependent on the relative motor speeds rather than absolute speeds. Equal speeds can yield stationary wheels, while slight speed variations between motors allow wheels to rotate either forward or backward, providing full bidirectional ground mobility. A schematic of the proposed architecture is shown in Figure~\ref{fig:both_sides}

Furthermore, most drone motors are sensorless brushless DC types due to their simplicity and cost efficiency, but this inherently leads to poor startup behavior at very low speeds. A differential mechanism can alleviate this problem by enabling both motors to run at a nominal idle speed during ground operation. In this scenario, as long as motor speeds remain matched, the wheel does not rotate, effectively bypassing the problematic motor startup region. Adjusting relative motor speeds slightly from this nominal point can then produce smooth and controlled wheel motion, improving ground-mode drivability without the need for additional actuators or complex mechanical switching.

\subsection{Experimental Vehicle Design}
The hybrid UAV prototype was developed as a custom quadrotor platform that integrates both aerial and ground mobility into a single compact vehicle. The design is organized into two distinct parts: the aerial side (top side) and the ground side (bottom side).

The top side of the vehicle closely resembles a conventional quadcopter. It has four A2212 1400kv BLDC motors mounted symmetrically on a lightweight 3D-printed frame. Each motor drives an 8-inch (8045) propeller that has a one-way bearing press-fitted and glued into its mounting hole. A shaft collar is installed on the top of each motor shaft to constrain the propeller. In flight mode, the motors spin forward, and the one-way bearings lock the propellers, allowing them to generate thrust. When the motors reverse for ground operation, the bearings allow the propellers to freewheel, effectively disengaging them so that energy is not wasted through propeller drag.

The top side of the vehicle houses the ground drivetrain. For ground propulsion, each pair of opposing motors transmits torque via small pulleys attached to their extended shafts. These pulleys drive belts that connect to larger pulleys on the input shafts of miniature differentials. Each differential then drives an integrated ground-contact wheel directly. The differential mechanism converts the relative speeds of the paired motors into bidirectional wheel rotation, enabling the vehicle to move forward, reverse, or execute turning maneuvers without additional actuators. To provide ground stability, two universal ball casters are mounted at the front and rear of the chassis.

This integrated configuration allows a quick transition between flight and ground modes through software-controlled reversal of motor rotation. In forward rotation (flight mode), the propellers are engaged to generate thrust, while in reverse rotation (ground mode) the propellers freewheel and the drivetrain—comprising the pulleys, belts, and differentials—provides efficient and bidirectional ground traction without extra actuators. The entire vehicle weighs only about 800\,g, which is notably light for a system of this scale incorporating ground mobility mechanisms.

\section{Control}
\label{sec:control}
\begin{figure}[h!]
  \centering
  \includegraphics[width=\linewidth]{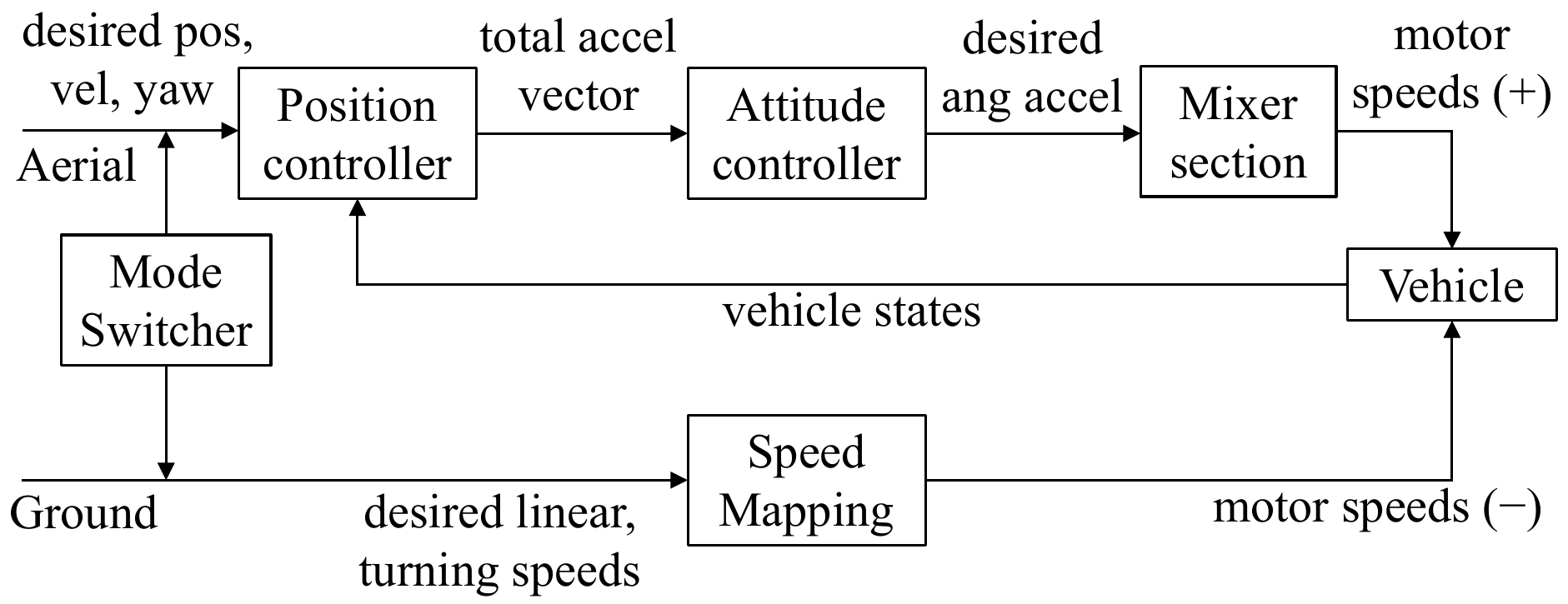}
  \caption{The diagram of the two-mode control framework. In aerial mode, a closed-loop cascaded controller manages acceleration, with inputs coming from either manual controls (for outdoor FPV flights) or an outer-loop position controller (for indoor Motion Capture flight). In ground mode, desired linear and rotational speeds are directly mapped to motor speed commands using factors such as gear ratio and wheel radius.}
  \label{fig:control_block}
\end{figure}

The vehicle uses a mode-switched control framework (Figure~\ref{fig:control_block}), allowing the same four motors to power either the propellers for flight or the ground wheels for driving. Mechanically, the belt-driven differentials remain engaged at all times, but the propellers are decoupled by one-way bearings when motors spin in reverse.

\subsection{Flight Mode}
In flight mode, the motors rotate forward and the one-way bearings allow the motor shafts to transmit torque to the propellers. Similar to many existing quadcopters, we employ a model-based cascaded controller architecture for Duawlfin in the flight mode. The controller enables autonomous tracking of desired position, velocity, and yaw. The controller can be decomposed into a position controller, an attitude controller, and a mixer section.

\subsubsection{Position Controller}
The position controller, in the outermost control loop, uses the desired and current positions and velocities to compute the desired acceleration. It then adds gravitational acceleration ($\boldsymbol{g}$) to produce the total acceleration needed to follow the trajectory.
\begin{equation}
    \begin{aligned}
    \boldsymbol{a}_{des} &= \omega_{nat}^2 (\boldsymbol{p}_{des} - \boldsymbol{p}_{cur}) + 2 \zeta \omega_{nat} (\boldsymbol{v}_{des} - \boldsymbol{v}_{cur}) \\
    \boldsymbol{a}_{tot} &= \boldsymbol{a}_{des} + \boldsymbol{g}
    \end{aligned}
\end{equation}

\subsubsection{Attitude Controller}
The attitude controller aligns the thrust direction with the desired total acceleration vector and applies the desired yaw. It computes the desired orientation, then derives the desired angular velocity and desired angular acceleration using current state feedback. The control law is as follows:
\begin{align}
\boldsymbol{\hat{a}}_{tot} &= \frac{\boldsymbol{a}_{tot}}{\| \boldsymbol{a}_{tot} \|} \\
\boldsymbol{n}_f &= \boldsymbol{\hat{z}} \times \boldsymbol{\hat{a}}_{tot} \\
\beta_f &= \arccos(\boldsymbol{\hat{a}}_{tot} \cdot \boldsymbol{\hat{z}}) \\
\boldsymbol{R}_f &= \mathcal{R_M}([\boldsymbol{n}_f/\|\boldsymbol{n}_f\|, \beta_f]) \\
\boldsymbol{R}_{des} &= \boldsymbol{R}_f \begin{bmatrix}
\cos\psi_{des} & -\sin\psi_{des} & 0 \\
\sin\psi_{des} & \cos\psi_{des} & 0 \\
0 & 0 & 1
\end{bmatrix} \\
\boldsymbol{\omega}_{des} &= \boldsymbol{\tau}_{att} \oslash \mathcal{V}(\boldsymbol{R}_{cur}^T \boldsymbol{R}_{des}) \\
\boldsymbol{e}_{\omega} &= \boldsymbol{\omega}_{des} - \boldsymbol{\omega}_{cur} \\
\boldsymbol{\alpha}_{des} &= \boldsymbol{\tau}_{\omega} \oslash \boldsymbol{e}_{\omega}
\end{align}

Where $\mathcal{R_M}$ is a function that converts an axis-angle representation to rotation matrix form, and $\mathcal{V}$ is a function that converts a rotation matrix to a 3D rotation vector.

\subsubsection{Mixer Section}
Given the vehicle's geometry, mass, moment of inertia, and motor constants, the mixer computes the individual rotor thrusts based on the total desired thrust and body torques. The outputs are then converted into motor speeds using propeller characteristics and sent as DShot signals to the ESC. Key controller parameters are summarized in Table \ref{controller}.
\begin{align}
f_{tot} &= m \| \boldsymbol{a}_{tot} \| \\
\boldsymbol{\tau} &= \boldsymbol{J}\boldsymbol{\alpha}_{des} \\
\begin{bmatrix}
f_1 \\
f_2 \\
f_3 \\
f_4
\end{bmatrix}
&=
\frac{1}{4}
\begin{bmatrix}
-1/l & -1/l & -1/k & 1 \\
-1/l &  1/l &  1/k & 1 \\
 1/l &  1/l & -1/k & 1 \\
 1/l & -1/l &  1/k & 1
\end{bmatrix}
\begin{bmatrix}
\tau_x \\
\tau_y \\
\tau_z \\
f_{tot}
\end{bmatrix}
\end{align}

\begin{table}[h!]
\centering
\caption{Vehicle controller parameters.}
\label{controller}
\begin{tabular}{c|c|c} 
\toprule
Symbol & Parameter & Value \\
\midrule
$\omega_{nat}$ & Natural frequency & 2 Hz \\
$\zeta$ & Damping ratio & 1 \\
$\boldsymbol{\tau}_{att}$ & Attitude time constant & [0.2, 0.2, 0.5] s \\
$\boldsymbol{\tau}_{\omega}$ & Rates time constant & [0.05, 0.05, 0.2] s \\
$l$ & Adjacent rotor spacing & 0.22 m \\
$k$ & Propeller thrust to torque constant & 0.014 m \\
\bottomrule
\end{tabular}
\end{table}

\subsection{Ground Mode}
In ground mode, the UAV’s desired linear speed $v_x$ (m/s) and angular speed $\omega_z$ (rad/s) from user inputs are used to compute the individual wheel speeds. The vehicle’s right and left wheel speeds, $w_r$ and $w_l$ respectively, are defined as:
\begin{equation}
    w_r = k_1\Bigl(v_x + k_2\,\omega_z\Bigr), \quad
    w_l = k_1\Bigl(v_x - k_2\,\omega_z\Bigr)
    \label{eq:wheel_speeds}
\end{equation}

where $k_1$ is the inverse of the wheel radius and $k_2$ is the lateral offset of the wheels from the vehicle center. Motor commands for ground mode are derived from the calculated wheel speeds. Let $w_i$ denote the idle motor speed, and let $\omega_1, \omega_2, \omega_3,$ and $\omega_4$ represent the commanded motor speeds for the four motors in their reverse directions. These are computed as follows:
\begin{equation}
    \begin{aligned}
    \omega_1 &= -k_3\,w_r + w_i, \quad
    \omega_2 = k_3\,w_r + w_i,\\
    \omega_3 &= k_3\,w_l + w_i, \quad
    \omega_4 = -k_3\,w_l + w_i
    \end{aligned}
    \label{eq:motor_commands}
\end{equation}

where $w_r$ and $w_l$ are the right and left wheel speeds defined in Equation~\eqref{eq:wheel_speeds}, and $k_3$ is the reduction ratio of the pulley-belt transmission. For robust ground operation, the motors are maintained above the nominal idle speed, and the minimum commanded speed is defined as
\begin{equation}
    \omega_{\min} = w_i
    \label{eq:omega_min}
\end{equation}

If any computed motor command $\omega_i$ falls below $\omega_{\min}$, that command is raised to $\omega_{\min}$ and the paired motor’s command is increased by the same amount to maintain the required speed differential. For example, if for a pair of motors the computed commands are $\omega_1$ and $\omega_2$ and $\omega_1 < \omega_{\min}$, then the adjustments are performed as follows:
\begin{equation}
\omega_1 \leftarrow \omega_{\min} \quad \text{and} \quad \omega_2 \leftarrow \omega_2 + \Bigl(\omega_{\min} - \omega_1\Bigr)
\end{equation}

This ensures that the intended speed difference $\omega_2 - \omega_1$ is maintained while avoiding operation in the low-performance startup region of sensorless brushless motors. 

\begin{figure}[tb]
\centering
\includegraphics[width=\linewidth]{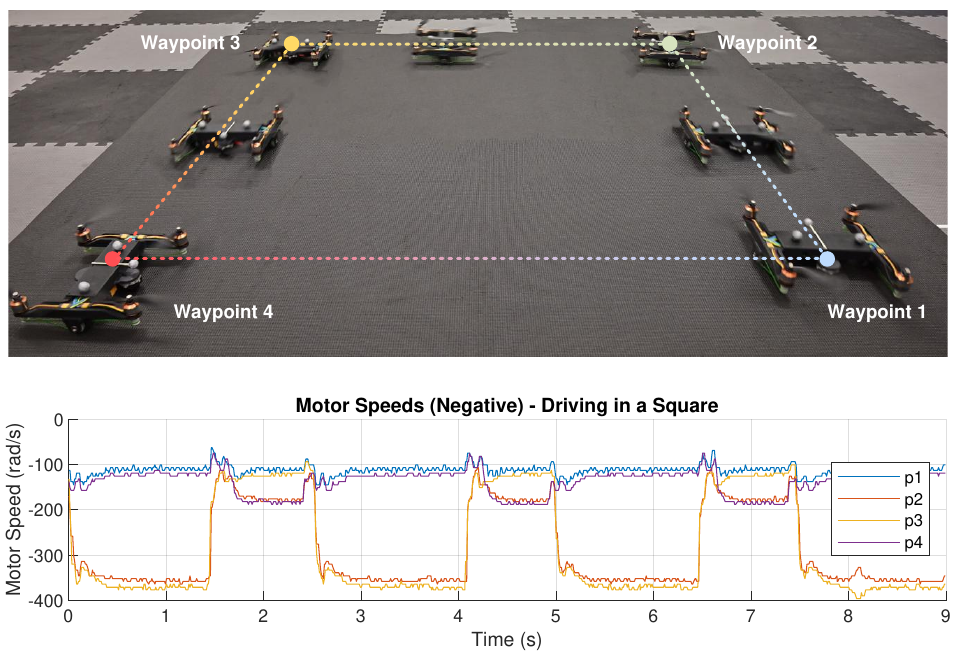}
\caption{Motor speed profiles for a complete square-path maneuver in the ground mode. The vehicle departs waypoint 1 and reaches waypoint 2 at $t = 1.5$~s, then performs a $90^\circ$ in-place rotation from $t = 1.5$ s to $2.5$~s. It travels to waypoint 3, arriving at $t = 4.0$~s, executes a second $90^\circ$ turn from $t = 4.0$~s to $5.0$~s, then moves to waypoint 4, arriving at $t = 6.5$~s. At waypoint 4 it performs a third $90^\circ$ rotation between $t = 6.5$~s and $7.5$~s before returning toward waypoint 1 to complete the square.}
\label{fig:square}
\end{figure}

To illustrate ground-mode behavior, we plot motor speeds for a simple square-path driving test. As shown in Figure~\ref{fig:square}, when the vehicle is moving in a straight line between every two waypoints, the rear motors on each side spin faster than the front motors, which remain at the minimum speed $\omega_i$: $|\omega_2| > |\omega_1| = |\omega_i|$ and $|\omega_3| > |\omega_4| = |\omega_i|$, causing both wheels to move forward. When the vehicle is turning around itself at a waypoint, the speed relationship becomes $|\omega_2| > |\omega_1| = |\omega_i|$ and $|\omega_4| > |\omega_3| = |\omega_i|$, so the right wheel continues to advance while the left wheel reverses, making the vehicle turn counterclockwise.

Mode switching does not require any physical reconfiguration beyond reversing motor spin directions. The high-level autonomy stack (or a user command) selects flight or drive mode, causing the controller to generate motor speed setpoints for either aerial thrust (forward rotation) or ground travel (reverse rotation). This unified approach enables rapid transitions with minimal mechanical complexity while preserving the vehicle's standard quadrotor functionality in flight.

\section{Experimental Validation}
\label{sec:exp}
The proposed hybrid UAV was evaluated in four sets of experiments that test its performance in both flight and ground modes, as well as its ability to transition seamlessly between these modes. Below, we present the key experiments along with discussions that highlight the performance trade-offs, advantages, and implications of the design.

\begin{figure}[h!]
        \centering
    \includegraphics[width=\linewidth]{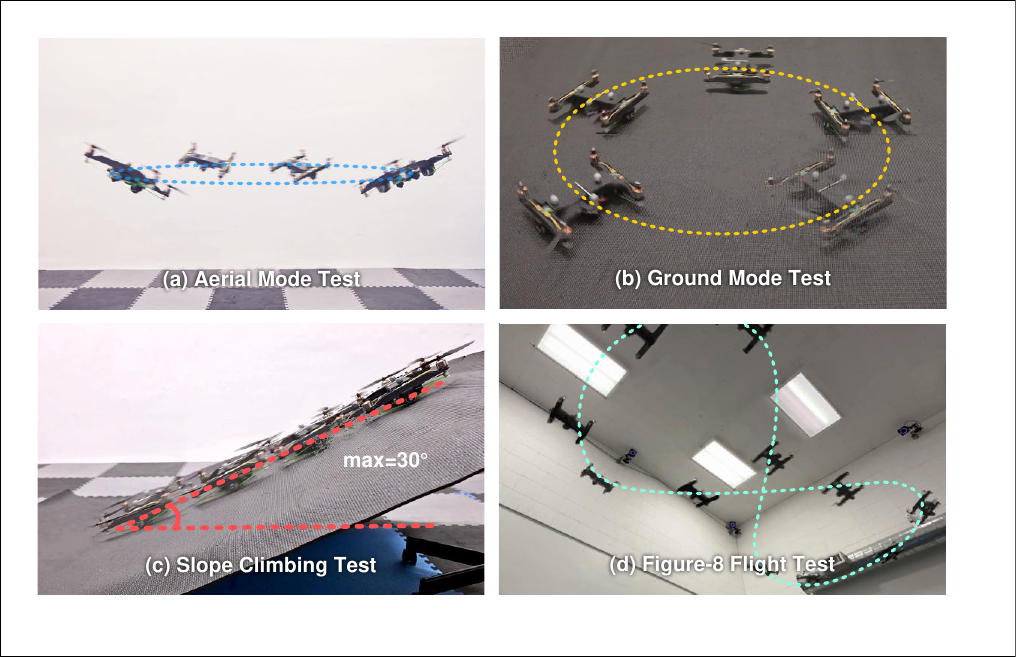}
    \caption{Time-lapse pictures of experiments.
    To test the power consumption and lateral acceleration under two different modes, we test and record the data when (a) Duawlfin flying in a circle. (b) Duawlfin driving in a circle. To explore the maximum slop-climbing capability, we let (c) Duawlfin drives up a ramp up to $30^\circ$. To test the performance of proposed controller, we make (d) Duawlfin fly along the figure-8 trajectory.}
    \label{fig:experiments}
\end{figure}

\subsection{Circular Motion Experiment}
\begin{table}[h!]
    \centering
    \scriptsize
    \caption{Average power consumption during circular motion (1-meter diameter) at various speeds in flight and ground modes.}
    \label{tab:circular_power}
    \begin{tabular}{c|c|c|c}
    \toprule
    Speed (m/s) & Lateral G-force (g) & Fly Power (W) & Drive Power (W) \\ \midrule
    1.0 & 0.20 & 124.6 & 3.9 \\ \hline
    1.5 & 0.46 & 140.6 & 8.2 \\ \hline
    2.0 & 0.82 & 188.8 & 14.9 \\ 
    \bottomrule
    \end{tabular}
\end{table}
To evaluate maneuverability and efficiency during tight turning motions, circular paths with a radius of 1 meter were tested in both flight and ground modes at speeds ranging from $1.0$ to $2.0$ m/s. Figure~\ref{fig:experiments}.(a) and (b) show the experiment footage. Table~\ref{tab:circular_power} summarizes the average power consumption for $5$ seconds and the lateral acceleration experienced in each scenario.

The results clearly demonstrate that for rapid turning maneuvers, the ground mode significantly outperforms flight mode in terms of energy efficiency. At $1.0$~m/s, ground-mode operation consumes only $3.9$ W, whereas flight mode requires $124.6$~W, which is over 30 times higher. Even at increased speeds of up to $2.0$~m/s, ground mode maintains drastically lower power demands compared to flight mode ($14.9$~W versus $188.8$~W). This substantial difference arises primarily because ground mode leverages tire-ground frictional forces, significantly reducing the power needed for directional control compared to aerodynamic maneuvering in flight.

Additionally, ground-mode operation exhibits exceptional agility, comfortably handling maneuvers that generate lateral accelerations approaching $1$~g ($0.82$~g at $2.0$~m/s). This indicates that the hybrid vehicle is not only energy-efficient but also highly capable of rapid directional changes in confined spaces, making it especially suitable for dynamic urban and indoor environments where tight maneuvering is frequently required.

In summary, the data confirms that ground-mode operation provides a marked advantage in energy efficiency and agility for short-range, rapid-turning maneuvers, enhancing the overall practicality and versatility of the hybrid UAV in realistic deployment scenarios.

\subsection{Slope Climbing Ability}
\begin{table}[h!]
    \centering
    \caption{Measured energy consumption per unit height climbed on different slope angles.}
    \label{tab:slope_energy}
    \begin{tabular}{c|c|c}
    \toprule
    Slope Angle (\textdegree) & Mean Power (W) & Energy / Height (J/m) \\ 
    \midrule
    5 & 4.0 & 50.9 \\ \hline
    10 & 4.6 & 37.9 \\ \hline
    15 & 4.5 & 27.5 \\ \hline
    20 & 4.5 & 24.8 \\ \hline
    25 & 5.0 & 25.9 \\ \hline
    30 & 5.4 & 26.5 \\ 
    \bottomrule
    \end{tabular}
\end{table}

To assess the ground-mode capability of the Duawlfin, climbing experiments as shown in Figure~\ref{fig:experiments}.(c) were conducted with slope inclines ranging from $5^\circ$ to $30^\circ$ at a fixed forward speed of $0.5$~m/s. The average power consumption and energy expenditure per unit height climbed were measured and summarized in Table \ref{tab:slope_energy}.

The experimental results highlight several important aspects of Duawlfin's ground performance. First, the power requirements for slope climbing remain consistently low, averaging between $4.0$~W to $5.4$~W across all tested angles, demonstrating high drivetrain efficiency. The data indicate that climbing is inherently energy-efficient, particularly on steeper slopes where the energy required per unit height actually decreases due to reduced traversal distance, reaching as low as $24.8$~J/m at $20^\circ$. This trend shows the system's effectiveness at converting energy into vertical displacement.

Additionally, the vehicle demonstrates robust climbing capabilities, successfully managing steep inclines up to $30^\circ$, at which point tire slippage begins to occur. Notably, a $30^\circ$ incline already represents a significantly challenging gradient far exceeding typical accessibility ramps mandated by ADA standards (maximum $4.76^\circ$). Thus, the UAV's climbing performance comfortably surpasses real-world urban navigation requirements, including indoor environments and common building terrains. These findings confirm that the differential drivetrain provides consistent, efficient, and reliable performance even on steep slopes, making the UAV highly suitable for versatile deployment scenarios involving varied terrain conditions.

\subsection{Figure‐8 Flight Test}

\begin{table}[t]
    \centering
    \scriptsize
    \caption{Hover and Figure‐8 Flight Performance}
    \label{tab:fig8}
    \begin{tabular}{l|c|c|c}
    \toprule
    Vehicle     & Hover Power (W) & Fig.-8 RMSE (cm) & Fig.-8 Power (W) \\
    \midrule
    Baseline    & 120.57           & 10.93            & 134.19            \\
    Duawlfin    & 124.35           & 9.74              & 138.53            \\
    \bottomrule
    \end{tabular}
\end{table}

\begin{figure}[t]
    \centering
    \includegraphics[width=0.9\linewidth]{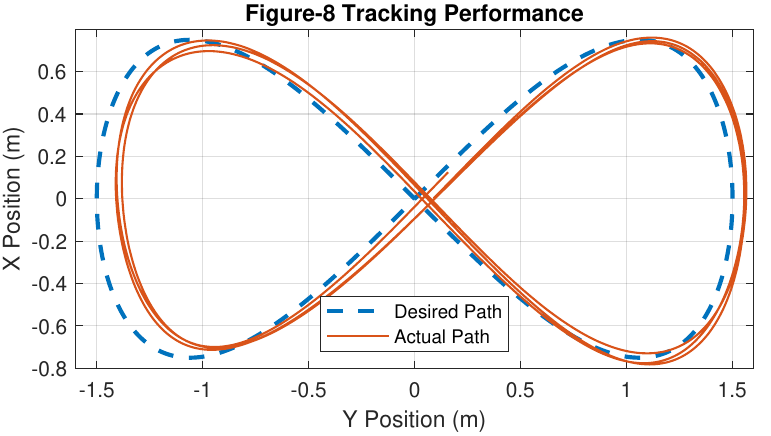}
    \caption{Figure‑8 flight test: the actual path (solid) of Duawlfin closely follows the desired trajectory (dashed), despite the additional inertial load of the ground drivetrain.}
    \label{fig:figure8}
\end{figure}
The addition of one‐way bearings and an integrated ground drivetrain alters Duawlfin’s rotor dynamics compared to a conventional quadcopter. Because the one‐way bearings block reverse torque transmission, active propeller braking is no longer possible. Moreover, the ground drivetrain remains engaged even in flight, so each rotor must accelerate the pulleys, belt, and differential—adding inertia and introducing frictional losses.

We argue that, despite these extra loads, the effects on performance are minimal. Propeller drag alone suffices to slow the rotors at an acceptable rate in the absence of active braking, as is common in many drones where the ESCs do not employ active braking. Additionally, the drivetrain components—such as plastic pulleys—are lightweight, resulting in only a small increase in total rotor inertia.

To investigate these effects, we executed a 3‐cycle figure-8 trajectory tracking flight on Duawlfin and a baseline vehicle. The figure‑8 trajectory is of $1.5$~m major diameter with the vehicle flying at a speed of $2$~m/s as shown in Figure~\ref{fig:experiments}.(d). The baseline vehicle is a modified Duawlfin with the same propellers but without the one‑way bearings and with its ground drivetrain disconnected; all other parameters remain identical.

Figure~\ref{fig:figure8} overlays the desired (dashed) and actual (solid) flight paths of Duawlfin, and Table~\ref{tab:fig8} summarizes the quantitative results of Duawlfin and the baseline vehicle in both hover flights and figure-8 trajectory tracking.

The data indicate that Duawlfin’s trajectory tracking root‑mean‑squared error (RMSE) is small and comparable to the baseline, demonstrating that the added drivetrain components have virtually no effect on tracking performance. Both hover power and figure‑8 trajectory tracking power are only about $3.23\%$ higher than those of the baseline, so the ground drivetrain imposes only a minor penalty on energy efficiency due to friction losses. These results confirm that the design modifications preserve robust aerial performance suitable for real‑world UAV applications.

In theory, active propeller braking and complete drivetrain disengagement in flight could still be implemented—e.g. either actively with servos or passively with clutches that engage when thrust is applied. However, the results presented here represent the most conservative case (drivetrain always engaged and no active braking), and even under these conditions the vehicle performs well.

\subsection{Outdoor Multi-Terrain Test}
\begin{figure}[t]
    \centering
    \includegraphics[width=\linewidth]{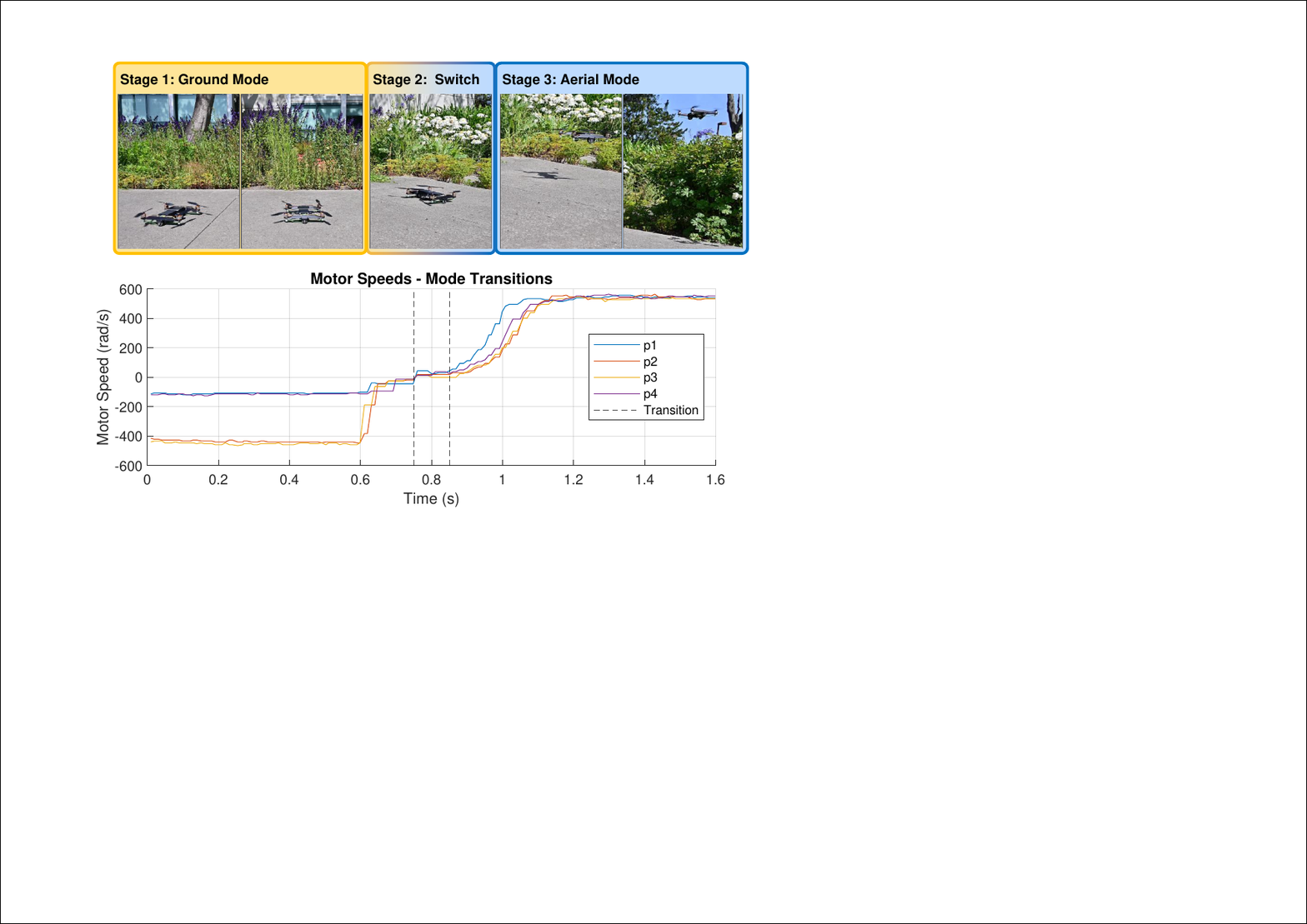}
    \caption{The demonstration of outdoor multi-terrain test.
    Stage 1: Duawlfin is running on the sidewalk. 
    Stage 2: Duawlfin switches from its ground mode into aerial mode. The mode switch is completed in 0.1s
    Stage 3: Duawlfin takes off and flies over the bush. With the complex terrain where the wheel locomotion cannot handle, the motors actuate propellers directly and enables the robot fly over it.
    }
    \label{fig:outdoor}
\end{figure}
We performed a multi-terrain demonstration to showcase Duawlfin's ability to transition between air and ground modes in a simulated delivery scenario. Specifically, we simulated a package delivery task from one building to another. For brevity and analytical clarity, only the initial phase—taking off and transitioning from ground to air outside the origin building—is presented here. Unlike all the other experiments, the flight mode controller in this part does not involve the position controller, and the vehicle is controlled manually with joystick-controlled acceleration instead. Figure~\ref{fig:outdoor} presents the results, with the top panel showing a time-lapse of the vehicle's path as it takes off from ground mode and transitions into flying, and the bottom panel displaying motor speeds and power consumption over time.

The vehicle data plot illustrates how quickly the vehicle can switch the operational mode by having the motors switch direction—from negative (driving the wheels) to positive (generating lift)—with minimal delay. This quick reversal of motor direction confirms that Duawlfin’s transition mechanism is both mechanically simple and dynamically responsive, enabling efficient deployment for delivery tasks requiring frequent mode changes.

Overall, this demonstration highlights the UAV's practical suitability for real-world urban logistics, showcasing its fast, stable, and energy-efficient transitions between aerial and ground operation modes.

\section{Conclusion}
\label{sec:conclusion}
We introduced Duawlfin, a drone with unified actuation for wheeled locomotion and flight operation that achieves efficient, bidirectional ground mobility without requiring additional actuators or relying on propeller thrust for terrestrial propulsion. The design employs one-way bearings and a differential system to repurpose existing quadrotor motors for both aerial flight and ground driving. This streamlined approach significantly reduces mechanical complexity, system weight, and energy usage while ensuring robust multimodal functionality.

Experimental validation confirmed the effectiveness and practicality of our hybrid robot across various scenarios. Flight tests indicated minimal performance trade-offs, with only minor increases in energy consumption compared to a conventional quadrotor in the flight mode. Ground-mode experiments demonstrated impressive efficiency, successfully navigating steep slopes of up to $30^\circ$ and performing agile turns with lateral accelerations close to $1$g, all while using significantly less energy than flying. The multi-terrain demo showcased the robot’s smooth and rapid transitions between aerial and terrestrial modes, underscoring its real-world applicability for urban delivery and indoor navigation tasks. 

Overall, Duawlfin provides a compelling alternative to existing hybrid UAV designs, offering substantial benefits in terms of efficiency, simplicity, and versatility. Future work will focus on further optimization of the mechanical drivetrain and integrating advanced autonomy to fully exploit the hybrid vehicle's capabilities in complex operational environments.

\section*{Acknowledgment}
This work is supported by Hong Kong Center for Logistics
Robotics. The experimental testbed at the HiPeRLab is the result of the contributions of many people, a full list of which can be found at {\tt\small\url{https://hiperlab.berkeley.edu/members/}}.

\bibliographystyle{ieeetr}
\bibliography{main}

\begin{thebibliography}{10}

\bibitem{mark2022design}
M.~W. Mueller, S.~J. Lee, and R.~D'Andrea, ``Design and control of drones,'' {\em Annual Review of Control, Robotics, and Autonomous Systems}, vol.~5, no.~1, pp.~161--177, 2022.

\bibitem{thusoo2021quadrotors}
R.~Thusoo, S.~Jain, and S.~Bangia, ``Quadrotors in the present era: a review,'' {\em Information Technology in Industry}, vol.~9, no.~1, pp.~164--178, 2021.

\bibitem{samouh2020multimodal}
F.~Samouh, V.~Gluza, S.~Djavadian, S.~Meshkani, and B.~Farooq, ``Multimodal autonomous last-mile delivery system design and application,'' in {\em 2020 IEEE International Smart Cities Conference (ISC2)}, pp.~1--7, IEEE, 2020.

\bibitem{raivi2023drone}
A.~M. Raivi, S.~A. Huda, M.~M. Alam, and S.~Moh, ``Drone routing for drone-based delivery systems: A review of trajectory planning, charging, and security,'' {\em Sensors}, vol.~23, no.~3, p.~1463, 2023.

\bibitem{bridgelall2024remote}
R.~Bridgelall, T.~Askarzadeh, D.~Tolliver, M.-P. Consortium, {\em et~al.}, ``Remote sensing of multimodal transportation assets using drones,'' tech. rep., Mountain-Plains Consortium, 2024.

\bibitem{jain2020iros}
K.~P. Jain, J.~Tang, K.~Sreenath, and M.~W. Mueller, ``Staging energy sources to extend flight time of a multirotor uav,'' in {\em 2020 IEEE/RSJ International Conference on Intelligent Robots and Systems (IROS)}, pp.~1132--1139, 2020.

\bibitem{kalantari2020drivocopter}
A.~Kalantari, T.~Touma, L.~Kim, R.~Jitosho, K.~Strickland, B.~T. Lopez, and A.-A. Agha-Mohammadi, ``Drivocopter: A concept hybrid aerial/ground vehicle for long-endurance mobility,'' in {\em Proceedings of the 2020 IEEE Aerospace Conference}, pp.~1--10, 2020.

\bibitem{qin2020hybrid}
Y.~Qin, Y.~Li, X.~Wei, and F.~Zhang, ``Hybrid aerial-ground locomotion with a single passive wheel,'' in {\em 2020 IEEE/RSJ International Conference on Intelligent Robots and Systems (IROS)}, pp.~1371--1376, 2020.

\bibitem{sugihara2024design}
K.~Sugihara, M.~Zhao, T.~Nishio, K.~Okada, and M.~Inaba, ``Design and control of delta: Deformable multilinked multirotor with rolling locomotion ability in terrestrial domain,'' {\em arXiv preprint arXiv:2403.06636}, 2024.

\bibitem{page2014quadroller}
J.~R. Page and P.~E.~I. Pounds, ``The quadroller: Modeling of a uav/ugv hybrid quadrotor,'' in {\em 2014 IEEE/RSJ International Conference on Intelligent Robots and Systems}, pp.~4834--4841, 2014.

\bibitem{sabet2019rollocopter}
S.~Sabet, A.-A. Agha-Mohammadi, A.~Tagliabue, D.~S. Elliott, and P.~E. Nikravesh, ``Rollocopter: An energy-aware hybrid aerial-ground mobility for extreme terrains,'' in {\em 2019 IEEE Aerospace Conference}, pp.~1--8, 2019.

\bibitem{premachandra2019study}
C.~Premachandra, M.~Otsuka, R.~Gohara, T.~Ninomiya, and K.~Kato, ``A study on development of a hybrid aerial / terrestrial robot system for avoiding ground obstacles by flight,'' {\em IEEE/CAA Journal of Automatica Sinica}, vol.~6, no.~1, pp.~327--336, 2019.

\bibitem{neng2023skywalker}
N.~Pan, J.~Jiang, R.~Zhang, C.~Xu, and F.~Gao, ``Skywalker: A compact and agile air-ground omnidirectional vehicle,'' {\em IEEE Robotics and Automation Letters}, vol.~8, no.~5, pp.~2534--2541, 2023.

\bibitem{zhang2023control}
R.~Zhang, J.~Lin, Y.~Wu, Y.~Gao, C.~Wang, C.~Xu, Y.~Cao, and F.~Gao, ``Model-based planning and control for terrestrial-aerial bimodal vehicles with passive wheels,'' in {\em 2023 IEEE/RSJ International Conference on Intelligent Robots and Systems (IROS)}, pp.~1070--1077, 2023.

\bibitem{lin2024skater}
J.~Lin, R.~Zhang, N.~Pan, C.~Xu, and F.~Gao, ``Skater: A novel bi-modal bi-copter robot for adaptive locomotion in air and diverse terrain,'' {\em IEEE Robotics and Automation Letters}, vol.~9, no.~7, pp.~6392--6399, 2024.

\bibitem{saemi2024heat}
F.~Saemi, A.~Whitson, and M.~Benedict, ``Heat transfer models and measurements of brushless dc motors for small uass,'' {\em Aerospace}, vol.~11, no.~5, p.~401, 2024.

\bibitem{morton2017mobile}
S.~Morton and N.~Papanikolopoulos, ``A small hybrid ground-air vehicle concept,'' in {\em 2017 IEEE/RSJ International Conference on Intelligent Robots and Systems (IROS)}, pp.~5149--5154, 2017.

\bibitem{zhao2023spider}
M.~Zhao, T.~Anzai, and T.~Nishio, ``Design, modeling, and control of a quadruped robot spidar: Spherically vectorable and distributed rotors assisted air-ground quadruped robot,'' {\em IEEE Robotics and Automation Letters}, vol.~8, no.~7, pp.~3923--3930, 2023.

\bibitem{sugihara2023flyhuman}
K.~Sugihara, M.~Zhao, T.~Nishio, T.~Makabe, K.~Okada, and M.~Inaba, ``Design and control of a small humanoid equipped with flight unit and wheels for multimodal locomotion,'' {\em IEEE Robotics and Automation Letters}, vol.~8, no.~9, pp.~5608--5615, 2023.

\bibitem{rs2018multi}
A.~R~S and m.~m. Dharmana, ``Multi-terrain multi-utility robot,'' {\em Procedia Computer Science}, vol.~133, pp.~651--659, 01 2018.

\bibitem{sihite2023m4}
E.~Sihite, A.~Kalantari, R.~Nemovi, A.~Ramezani, and M.~Gharib, ``Multi-modal mobility morphobot (m4) with appendage repurposing for locomotion plasticity enhancement,'' {\em Nature communications}, vol.~14, no.~1, p.~3323, 2023.

\end{thebibliography}
\balance
\end{document}